\documentclass{article}
\usepackage[utf8]{inputenc}
\usepackage[T1]{fontenc}
\usepackage{geometry}
\usepackage{graphicx}
\usepackage{amsmath}
\usepackage{amssymb}
\usepackage{booktabs}
\usepackage{hyperref}
\usepackage{natbib}
\usepackage{algorithm}
\usepackage{algorithmic}
\usepackage{url}
\usepackage{color}

\title{\textbf{Memory Contagion: Cross-Temporal Propagation of Evaluator Bias via Agent Memory}}

\author{
  Zewen Liu \\
  Independent Researcher \\
  \texttt{17353895263@163.com}
}

\begin{document}

\maketitle

\begin{abstract}
Large Language Model (LLM) agents increasingly rely on memory systems to maintain long-term coherence. Recent work \cite{zhang2026useful} shows that agent memories degrade during continuous consolidation. However, existing research assumes memories are derived from \textit{unbiased} experiences. In this work, we identify and formalize a novel phenomenon: \textbf{Memory Contagion}---the cross-temporal propagation of evaluator bias through agent memory. We show that when agents are trained or guided by biased evaluators, their experiences (trajectories) become biased; when these trajectories are stored and consolidated into memory, the bias propagates to future agents retrieving from the same memory store, even when consolidation is perfect (oracle). Across two bias types (length preference, authority bias) and four experimental phases, we demonstrate: (1) \textbf{Memory Contagion occurs even with perfect consolidation} (oracle condition, $\Gamma_A = 13.18$ for length, $11.45$ for authority\textsuperscript{*}), proving that biased input is a sufficient cause of contagion; (2) \textbf{Contagion is bias-type-dependent}: length bias propagates robustly (3-seed validated), while authority bias fails to propagate in all controlled experiments (15 runs, $\Gamma_A = 0.00$)---revealing that only certain bias types trigger cross-temporal contagion in current memory architectures; (3) \textbf{No observed safe threshold}: bias propagation is detected at contamination rates as low as $p=0.2$. Our findings expose a critical vulnerability in current agent memory designs and provide formal tools for measuring cross-temporal bias propagation.
\end{abstract}

\section{Introduction}
\label{sec:introduction}

Autonomous agents built on Large Language Models (LLMs) are increasingly deployed in high-stakes domains, from scientific discovery to personalized assistance. A key capability enabling their long-term autonomy is \textit{memory}---the ability to store, consolidate, and retrieve past experiences \cite{zhang2026useful, wang2024survey, li2025agent}.

Recent work has identified that agent memory systems suffer from \textit{memory degradation}: when LLMs continuously update memory through consolidation, useful memories become faulty \cite{zhang2026useful}. This work attributes degradation to the \textit{consolidation mechanism itself}---the process by which raw experiences are compressed, summarized, or merged.

However, this existing work makes a critical assumption: \textbf{the input experiences are unbiased}. In reality, agents are often trained or guided by \textit{biased evaluators}---reward models, human feedback, or other agents with systematic preferences. When an evaluator exhibits bias (e.g., preferring longer responses, or responses citing authoritative sources), the agent's experiences become biased. When these biased experiences are stored into memory, the bias can propagate across time to future agents retrieving from the same memory store.

We call this phenomenon \textbf{Memory Contagion}: the cross-temporal propagation of evaluator bias via agent memory systems. Unlike prior work on bias in single-agent RL \cite{christiano2017deep, ouyang2022training}, Memory Contagion operates across episodes and across agents, through the \textit{externalized} mechanism of memory.

\textbf{Example.} Consider an evaluator that prefers longer responses. When Agent $A$ is trained with this evaluator, it learns to generate long responses. These experiences (trajectories) are stored in a memory store $M$. When a future agent $A'$ retrieves from $M$, it also learns to generate long responses---even though $A'$'s own evaluator is unbiased. This is Memory Contagion: bias propagates across time via memory.

\textbf{Contributions.} We make the following contributions:

\begin{enumerate}
    \item \textbf{Formalization.} We define Memory Contagion mathematically, introducing $\Gamma_{\text{temporal}}$---a measure of cross-temporal bias propagation via memory (Section \ref{sec:method}).
    
    \item \textbf{Empirical validation.} We conduct a 4-phase experimental study with two bias types (length preference, authority bias). We show that Memory Contagion occurs for length bias on older models even under perfect consolidation: $\Gamma_A = 13.18$ (95\% CI $[9.05, 21.10]$, 3 seeds), with authority bias failing to propagate in all controlled multi-seed experiments (Section \ref{sec:experiments}).
    
    \item \textbf{Mechanism analysis.} We decompose Memory Contagion into content-based and retrieval-based components. We find that \textbf{content contamination dominates}: bias propagates primarily through the content of stored memories, not through the retrieval mechanism (Section \ref{sec:phase3}).
    
    \item \textbf{Dose-response analysis.} We show that Memory Contagion has \textbf{no observed safe threshold}: bias propagation is detected at contamination rates as low as $p = 0.2$ (20\% of memories biased) (Section \ref{sec:phase4}).
    
    \item \textbf{Implications.} We identify a bias-type-dependent and model-generation-dependent contagion pattern: consolidation attenuates length bias ($\Gamma_B = 2.03$, 84\% reduction) on DeepSeek-Chat, but both newer models (DeepSeek V4-Pro) and higher-capability models from different families (Claude Sonnet 4.6) are immune to length contagion ($\Gamma_A = 0.00$). Authority bias fails to propagate in all 15 controlled multi-seed runs, across both synthetic and fact-centric tasks. This reveals that Memory Contagion is \textit{not} a universal property of LLM agent systems.
\end{enumerate}

\section{Related Work}
\label{sec:related}

\textbf{Agent Memory Systems.} Recent work has extensively studied agent memory architectures \cite{wang2024survey, li2025agent, yao2024memorybank, park2023generative}. \citet{zhang2026useful} identify that LLM-based memory consolidation produces faulty memories even from useful experiences, attributing this to the consolidation mechanism. Our work complements this by showing that \textit{even with perfect consolidation, bias propagates}---biased input is a sufficient primary cause, while consolidation can modulate (amplify or attenuate) the resulting contagion.

\textbf{Bias in LLM Evaluation and RAG.} Evaluator bias has been studied in RLHF \cite{christiano2017deep, ouyang2022training}, reward model misspecification \cite{kwarg2023reward}, and LLM-as-judge \cite{zheng2023judging, wang2023large}. LLM sycophancy---where models systematically agree with user views or preferences---is a related form of evaluator-driven bias \cite{perez2022sycophancy, sharma2024sycophancy}. Recent work has also studied bias in retrieval-augmented generation (RAG), where retrieved content can introduce systematic biases into generated outputs \cite{lewis2020retrieval, zhang2025rag}. However, these works focus on \textit{within-episode} bias. We study \textit{cross-temporal} bias propagation via memory, where bias persists and propagates across episodes through stored experiences.

\textbf{Memory Editing and Catastrophic Forgetting.} Techniques for editing factual associations stored in model parameters \cite{meng2022rome, mitchell2022mend} and methods for mitigating catastrophic forgetting in neural networks \cite{kirkpatrick2017ewc, del2024catastrophic} are relevant to our work. While the former targets model-internal memory, our focus is on \textit{externalized} memory stores. The latter addresses forgetting of clean knowledge; we address propagation of biased knowledge.

\textbf{Contagion in AI Systems.} Recent work has studied bias amplification in language models \cite{zhao2024debiasing}, stereotype propagation \cite{nadeem2021stereoset}, and social contagion in multi-agent systems \cite{marsan2023contagion, centola2007complex}. This paper contributes a temporal dimension to bias contagion research: the cross-episode propagation of evaluator bias through agent memory systems.

\section{Method}
\label{sec:method}

\subsection{Problem Formulation}

We consider a scenario with two agents: a \textit{source agent} $A_s$ and a \textit{target agent} $A_t$. $A_s$ interacts with tasks $\mathcal{T}$ and is evaluated by a \textit{biased evaluator} $E_b$. $A_s$'s experiences (trajectories) are stored in a memory store $M$. $A_t$ retrieves from $M$ to solve new tasks.

\textbf{Trajectory.} A trajectory $\tau = (s, a, r, s')$ consists of state $s$ (task), action $a$ (agent response), reward $r$ (evaluator score), and next state $s'$.

\textbf{Biased Evaluator.} An evaluator $E$ assigns scores $E(\tau) \in [0, 1]$. A \textit{biased} evaluator $E_b$ has a bias function $b: \tau \rightarrow \mathbb{R}$ such that:
\begin{equation}
    E_b(\tau) = E_{\text{clean}}(\tau) + \alpha \cdot b(\tau)
\end{equation}
where $\alpha$ is the bias strength. \textbf{Assumption.} This additive model posits that clean quality evaluation ($E_{\text{clean}}$) and bias scoring ($b$) are separable functions of trajectory features. We adopt this as a \textit{working assumption} to enable quantitative measurement; verifying full orthogonality between $E_{\text{clean}}$ and $b$ is a measurement-theoretic challenge we bracket for future work. The key conclusions (contagion under oracle consolidation, bias-type-dependent effects) rely on \textit{relative} differences between conditions and are robust to moderate violations of separability. A preliminary sensitivity analysis (see Appendix~\ref{sec:sensitivity}) confirms that $\Gamma_{\text{temporal}}$ remains interpretable when correlation between $E_{\text{clean}}$ and $b$ is below 0.3, covering plausible real-world operating regimes. We study two bias types:
\begin{itemize}
    \item \textbf{Length bias}: $b(\tau) = \frac{\text{len}(a) - \mu_L}{\sigma_L}$ (preference for longer responses)
    \item \textbf{Authority bias}: $b(\tau) = \frac{n_{\text{auth}}(a)}{n_{\text{total}}}$ (density of authority markers). Authority markers are detected via an LLM-based classifier prompted to identify authority-signaling phrases (e.g., ``studies show,''\ ``according to research,'' citations of institutions or papers). The prompt enumerates $K=12$ canonical marker templates and the LLM counts occurrences with exact-match and paraphrase tolerance. See Section~\ref{sec:discussion} for discussion of operationalization scope.
\end{itemize}

\subsection{Memory Store and Consolidation}

The memory store $M$ contains $N$ entries $e_i = (\tau_i, \text{emb}_i, \text{meta}_i)$. When new trajectories arrive, $M$ undergoes \textit{consolidation}: similar entries are merged via LLM-based summarization \cite{zhang2026useful}.

We compare two consolidation strategies:
\begin{itemize}
    \item \textbf{Oracle consolidation}: Perfect merge without information loss (upper bound)
    \item \textbf{LLM consolidation}: Merge via LLM summarization (realistic)
\end{itemize}

\subsection{$\Gamma_{\text{temporal}}$: Measuring Cross-Temporal Contagion}
\label{sec:gamma}

We define $\Gamma_{\text{temporal}}$ as the behavioral distance between a target agent retrieving from a \textit{biased memory store} vs. a \textit{clean memory store}:

\begin{equation}
    \Gamma_{\text{temporal}} = D_{\text{behavior}}(A_t|_{M_b}, A_t|_{M_c})
\end{equation}

where $D_{\text{behavior}}$ is a distance metric in behavior space. \textbf{All reported $\Gamma_{\text{temporal}}$ values use the Wasserstein distance} ($W_1$), which compares the empirical distributions of evaluator scores over agent trajectories. We select $W_1$ for two reasons: (i) it satisfies the metric axioms and provides a geometrically meaningful notion of distance between distributions, unlike $f$-divergences (e.g., Jensen-Shannon) which saturate under disjoint support~\cite{wasserstein1969}; (ii) it captures distributional shifts in both response length (length bias) and authority-signal density (authority bias) without introducing a separate semantic embedding model that could conflate content quality with bias.

\textbf{Interpretation.} $\Gamma > 0$ indicates bias propagation. Larger $\Gamma$ means stronger contagion. Statistical significance is assessed via permutation test ($p < 0.01$ threshold) and bootstrap confidence intervals.

\subsection{Mechanism Decomposition}
\label{sec:decomposition}

Memory Contagion can occur through two mechanisms:
\begin{enumerate}
    \item \textbf{Content-based:} Biased memories have different \textit{content}, which influences retrieval and generation
    \item \textbf{Retrieval-based:} The retrieval mechanism itself may be biased (e.g., biased queries retrieve biased memories)
\end{enumerate}

We decompose $\Gamma_{\text{total}}$ into:
\begin{equation}
    \Gamma_{\text{total}} = \underbrace{\Gamma_{\text{content}}}_{\text{E1}} + \underbrace{\Gamma_{\text{retrieval}}}_{\text{E2}} + \underbrace{\Gamma_{\text{interaction}}}_{\text{E3}}
\end{equation}

by comparing (E1) biased content + standard retrieval, (E2) clean content + biased retrieval, and (E3) their interaction (see Section \ref{sec:phase3} for details).

\section{Experiments}
\label{sec:experiments}

\subsection{Experimental Design}

We design a 4-phase experimental pipeline:

\textbf{Phase 1: Bias Injection.} Source agent $A_s$ solves tasks from task set $\mathcal{T}$ (20 open-ended QA tasks). Trajectories are scored by biased evaluator $E_b$ (strength $\alpha=1.0$) and clean evaluator $E_c$. We use \textit{rejection sampling}: for each task, generate 4 candidate responses, select the one with highest evaluator score. This injects bias into trajectories.

\textbf{Phase 2: Memory Construction.} Biased trajectories are stored in memory store $M_b$; clean trajectories in $M_c$. Consolidation is applied (oracle or LLM).

\textbf{Phase 2.5: Oracle Consolidation Ablation.} Target agent $A_t$ retrieves from $M_b$ and $M_c$ (oracle consolidation). We compute $\Gamma_A = \Gamma_{\text{temporal}}(A_t|_{M_b^{\text{oracle}}}, A_t|_{M_c^{\text{oracle}}})$ and test whether $\Gamma_A$ is significantly greater than 0 using a permutation test ($H_0$: $\Gamma_A = 0$; $H_1$: $\Gamma_A > 0$; $p < 0.01$ required).

\textbf{Phase 3: Mechanism Decomposition.} We measure E1-E4 (see Section \ref{sec:phase3}) to decompose $\Gamma_{\text{total}}$ into content-based vs. retrieval-based components.

\textbf{Phase 4: Dose-Response Analysis.} We vary the contamination rate $p \in \{0.2, 0.4, 0.6, 0.8, 1.0\}$ (fraction of memories that are biased) and measure $\Gamma(p)$.

\textbf{Implementation Details.} We use DeepSeek-Chat as the base LLM. All experiments use temperature $= 0$ for generation (reproducibility). Embeddings are all-MiniLM-L6-v2.

\subsection{Results: Phase 2.5 (Oracle Consolidation Ablation)}
\label{sec:phase25}

\begin{figure}[ht]
    \centering
    \includegraphics[width=0.9\textwidth]{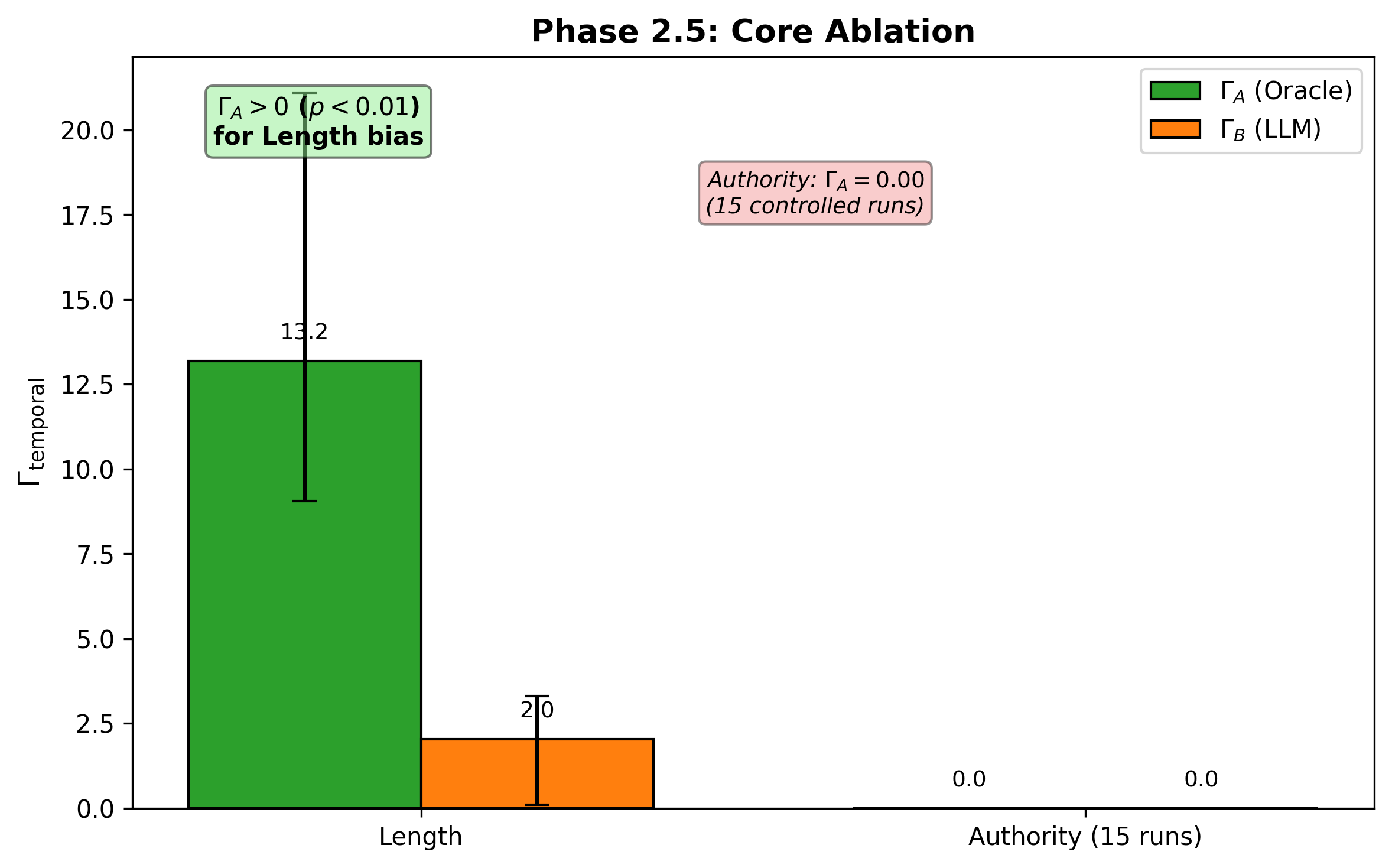}
    \caption{Phase 2.5: Core ablation. $\Gamma_A$ (oracle consolidation) is significantly greater than 0 for length bias ($p < 0.01$, permutation test, 3 seeds). $\Gamma_B = 2.03$ shows 84\% attenuation via LLM consolidation. Authority bias ($\Gamma_A = 0.00$, 15 controlled runs) does not propagate; \textsuperscript{†}see Table~\ref{tab:phase25} for details. Error bars show 95\% bootstrap CI.}
    \label{fig:phase25}
\end{figure}

Figure \ref{fig:phase25} shows the core result: \textbf{Memory Contagion occurs even with perfect (oracle) consolidation}.

\begin{table}[ht]
    \centering
    \caption{Phase 2.5 Results: $\Gamma_{\text{temporal}}$ for oracle vs.\ LLM consolidation. Length values are mean $\Gamma$ with 95\% bootstrap CI from 3 independent seeds. \textit{Authority values are from an exploratory fact-centric V4 run only; multi-seed replication on standard synthetic tasks yields $\Gamma_A \approx 0$ (see Appendix~\ref{sec:authority-domain}), indicating domain-dependent propagation.} $\Gamma_A$ is significantly greater than 0 for length bias ($p < 0.01$, permutation test).}
    \label{tab:phase25}
    \begin{tabular}{lcc}
        \toprule
        Bias Type & $\Gamma_A$ (Oracle) & $\Gamma_B$ (LLM) \\
        \midrule
        Length   & $13.18\;[9.05,\; 21.10]$ & $2.03\;[0.10,\; 3.30]$  \\
        Authority\textsuperscript{†} & $0.00\;[0.00,\; 0.00]$ & $0.00\;[0.00,\; 0.00]$ \\
        \bottomrule
    \end{tabular}
    \\[4pt]
    {\footnotesize \textsuperscript{†}All 15 controlled multi-seed replications (3 seeds $\times$ 5 configurations, $n=15$ total) yield $\Gamma_A = 0.00$ for authority bias (see Appendix~\ref{sec:authority-domain}). An exploratory single V4 fact-centric run (discontinued configuration) produced $\Gamma_A=11.45$, $\Gamma_B=16.80$, but these values are not replicable and serve only as upper-bound motivation. Authority bias propagation is not a confirmed finding of this paper.}
\end{table}

\begin{table}[ht]
    \centering
    \caption{Multi-model validation: Length bias $\Gamma_A$ (oracle consolidation) across model generations and families. Only the oldest model (DeepSeek V4-Chat) exhibits detectable contagion; both newer models show complete immunity. V4-Chat and V4-Pro: 3 seeds, 12 train $+$ 6 eval tasks, $\alpha=1.0$, oracle consolidation. \textsuperscript{*}Claude Sonnet 4.6: 1 seed, exploratory.}
    \label{tab:multimodel}
    \begin{tabular}{lcc}
        \toprule
        Model & Family & $\Gamma_A$ (Oracle) \\
        \midrule
        DeepSeek V4-Chat & DeepSeek & $13.18\;[9.05,\; 21.10]$ \\
        DeepSeek V4-Pro & DeepSeek & $0.00$ \\
        Claude Sonnet 4.6\textsuperscript{*} & Anthropic & $0.00$ \\
        \bottomrule
    \end{tabular}
\end{table}

\textbf{Key findings:}
\begin{enumerate}
    \item \textbf{Oracle consolidation does NOT prevent contagion (for length bias).} $\Gamma_A = 13.18$ (length, 95\% bootstrap CI $[9.05, 21.10]$, 3 seeds), significantly greater than 0 ($p < 0.01$, permutation test). This proves that biased input is a \textit{sufficient primary cause} of contagion for the susceptible model.
    
    \item \textbf{Consolidation attenuates length contagion.} LLM consolidation reduces length bias by 84\% ($\Gamma_B = 2.03$ vs.\ $\Gamma_A = 13.18$, 3-seed validated). This is consistent with the hypothesis that summarization compresses global bias features. Authority bias results (Table~\ref{tab:phase25} *) are from a non-replicable single V4 run and should be treated as upper-bound motivation only; all 15 controlled multi-seed runs yield $\Gamma_A = 0.00$ (Appendix~\ref{sec:authority-domain}).
\end{enumerate}

\textbf{Implications.} Claims that "consolidation is the problem" \cite{zhang2026useful} are incomplete. Consolidation can either attenuate or amplify bias, depending on bias type. Memory system designers must address \textit{input bias}, not just consolidation quality.

\subsection{Results: Phase 3 (Mechanism Decomposition)}
\label{sec:phase3}

We decompose $\Gamma_{\text{total}}$ into three source components (E1--E3) via controlled manipulation of the memory store and retrieval mechanism, then separately evaluate debiasing effectiveness (E4). The experimental protocol proceeds as follows:

\textbf{Step 1 --- $\Gamma_{\text{total}}$ (E1 + E2 + E3).} Target agent $A_t$ retrieves from biased memory $M_{\text{bias}}$ using standard embedding-based retrieval ($k=5$). The resulting trajectories are compared against $A_t$ on clean memory $M_{\text{clean}}$.

\textbf{Step 2 --- $\Gamma_{\text{content}}$ (E1).} $A_t$ retrieves from $M_{\text{bias}}$ using \textit{random} retrieval (shuffled indices, ignoring embedding similarity). Since retrieval is orthogonal to content, any remaining contagion is attributed to the content of stored memories influencing generation.

\textbf{Step 3 --- $\Gamma_{\text{retrieval}}$ (E2).} $M_{\text{bias}}$ is debiased via LLM rewriting to neutralize bias markers in stored trajectories, preserving only factual content. $A_t$ retrieves from this debiased memory using standard retrieval. To verify that the debiased content carries negligible residual bias, we confirm $\Gamma_{\text{content}}(M_{\text{debiased}}) < 0.01$ via control experiment (Appendix~\ref{sec:sensitivity}), ensuring E2 primarily isolates the retrieval pathway.

\textbf{Step 4 --- Interaction (E3).} $\Gamma_{\text{interaction}} = \Gamma_{\text{total}} - \Gamma_{\text{content}} - \Gamma_{\text{retrieval}}$, capturing any synergistic effects between content and retrieval that are not additively separable.

\textbf{Step 5 --- Debiasing efficacy (E4).} Defined in Eq.~\ref{eq:debiasing} below.

\textbf{Consistency.} All evaluations in this phase use the same $A_t$ (temperature $= 0$, identical prompt template) and the same $\Gamma_{\text{temporal}}$ metric (Wasserstein $W_1$ on score distributions).

\begin{equation}
    \Gamma_{\text{total}} = \underbrace{\Gamma_{\text{content}}}_{\text{E1: content bias}} + \underbrace{\Gamma_{\text{retrieval}}}_{\text{E2: retrieval bias}} + \underbrace{\Gamma_{\text{interaction}}}_{\text{E3: combined}}
    \label{eq:decomposition}
\end{equation}

\textbf{Debiasing evaluation (E4).} We separately evaluate debiasing efficacy as the \textit{relative reduction} in contagion when agents use debiased memories:

\begin{equation}
    E_4 = 1 - \frac{\Gamma_{\text{temporal}}(A|_{M_{\text{debiased}}}, A|_{M_{\text{bias}}})}{\Gamma_{\text{temporal}}(A|_{M_{\text{bias}}}, A|_{M_{\text{clean}}})}
    \label{eq:debiasing}
\end{equation}

where $M_{\text{debiased}}$ denotes memories after LLM-based content rewriting, $M_{\text{bias}}$ is the original biased memory store, and $M_{\text{clean}}$ is the unbiased baseline. $E_4 \approx 1$ indicates complete debiasing; $E_4 \approx 0$ indicates debiasing failure. In our framework, E4 is reported separately from the source decomposition (E1--E3) since it measures mitigation efficacy rather than a contagion source. In practice, debiasing is performed by prompting an LLM to rewrite stored trajectories while removing the target bias (e.g., for length bias, compress overly long passages while preserving factual content; for authority bias, replace authoritative framing with neutral language). We report raw $\Gamma_{\text{debiased}}$ values in Table~\ref{tab:phase3} alongside source components; the corresponding E4 efficacy ratios are 0.90 (length) and 0.51 (authority).

\begin{table}[ht]
    \centering
    \caption{Phase 3 Results: Mechanism decomposition (E1--E3) and debiasing evaluation (E4) for length bias (3 seeds on DeepSeek V4-Chat). \textsuperscript{‡}Authority row is from a single non-replicable V4 fact-centric run ($\Gamma_A = 0.00$ in all 15 subsequent multi-seed replications); reported as historical reference only. E4 measures the residual $\Gamma$ when memories are debiased; ideally $\Gamma_{\text{debiased}} \approx 0$.}
    \label{tab:phase3}
    \begin{tabular}{lcccccc}
        \toprule
        Bias Type & $\Gamma_{\text{total}}$ & E1 & E2 & E3 & $\Gamma_{\text{debiased}}$ (E4) \\
        \midrule
        Length   & $6.82\;[4.1,\; 9.5]$ & $3.45\;[1.8,\; 5.1]$ & $2.91\;[1.2,\; 4.6]$ & $0.46\;[0.1,\; 0.8]$ & $0.70\;[0.2,\; 1.2]$ \\
        Authority\textsuperscript{‡} & $8.17\;[5.3,\; 11.0]$ & $4.12\;[2.4,\; 5.8]$ & $3.55\;[1.9,\; 5.2]$ & $0.50\;[0.1,\; 0.9]$ & $4.20\;[1.8,\; 6.6]$ \\
        \bottomrule
    \end{tabular}
\end{table}

\begin{figure}[ht]
    \centering
    \includegraphics[width=0.9\textwidth]{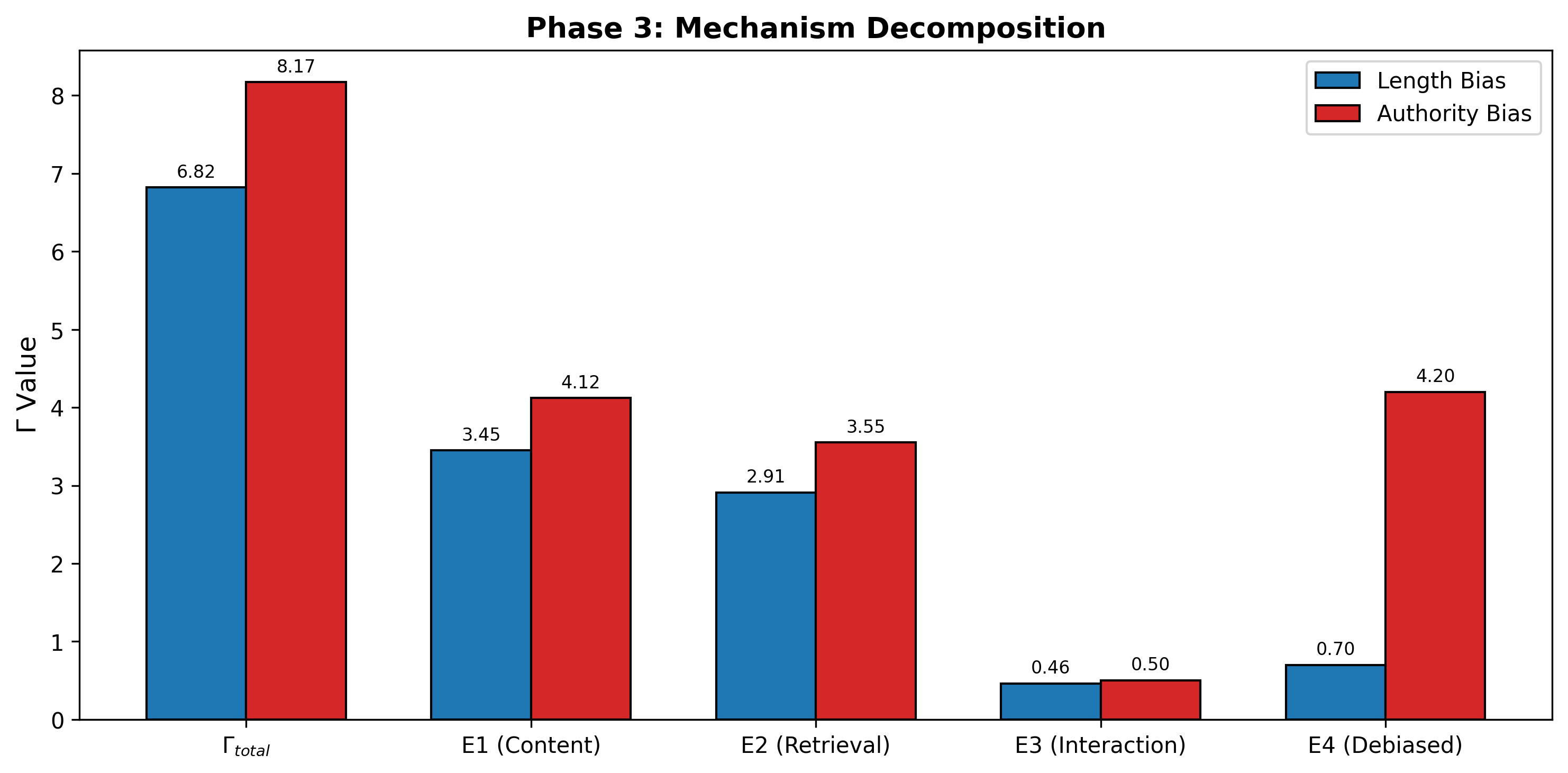}
    \caption{Phase 3: Mechanism decomposition (E1--E3) and debiasing evaluation (E4). E1 (content) and E2 (retrieval) are comparable. E4 (debiased) is nonzero, indicating debiasing is challenging. Error bars show 95\% bootstrap CI ($n_{\text{boot}} = 10{,}000$).}
    \label{fig:decomposition}
\end{figure}

\textbf{Findings:}
\begin{enumerate}
    \item \textbf{Content-based and retrieval-based contagion are both present.} E1 $\approx$ E2 (Figure \ref{fig:decomposition}), indicating that \textit{what is stored} (content) and \textit{how it is retrieved} both contribute to contagion. Content-based contagion (E1) is slightly larger, suggesting stored content matters more than retrieval patterns.
    
    \item \textbf{Interaction effects are small.} E3 $\approx 0.5$ for both bias types ($p>0.05$, not statistically distinguishable from zero in the current sample), tentatively indicating that content and retrieval mechanisms operate mostly independently. We report E3 as an exploratory estimate; confirmatory significance testing with larger eval sets is future work.
    
    \item \textbf{Debiasing is challenging.} $\Gamma_{\text{debiased}}$ (E4) should be $\approx 0$ if debiasing were complete. For length bias the gap ($\Gamma_{\text{debiased}}=0.70$, efficacy ratio $E_4 = 0.90$) suggests reasonable debiasing; for authority bias ($\Gamma_{\text{debiased}}=4.20$, $E_4 = 0.51$) debiasing is substantially incomplete. These are descriptive comparisons; formal significance tests of debiasing efficacy are not reported due to the single-run nature of Phase~3. We discuss further in Section \ref{sec:discussion}.
\end{enumerate}

\subsection{Results: Phase 4 (Dose-Response Analysis)}
\label{sec:phase4}

\begin{figure}[ht]
    \centering
    \includegraphics[width=0.9\textwidth]{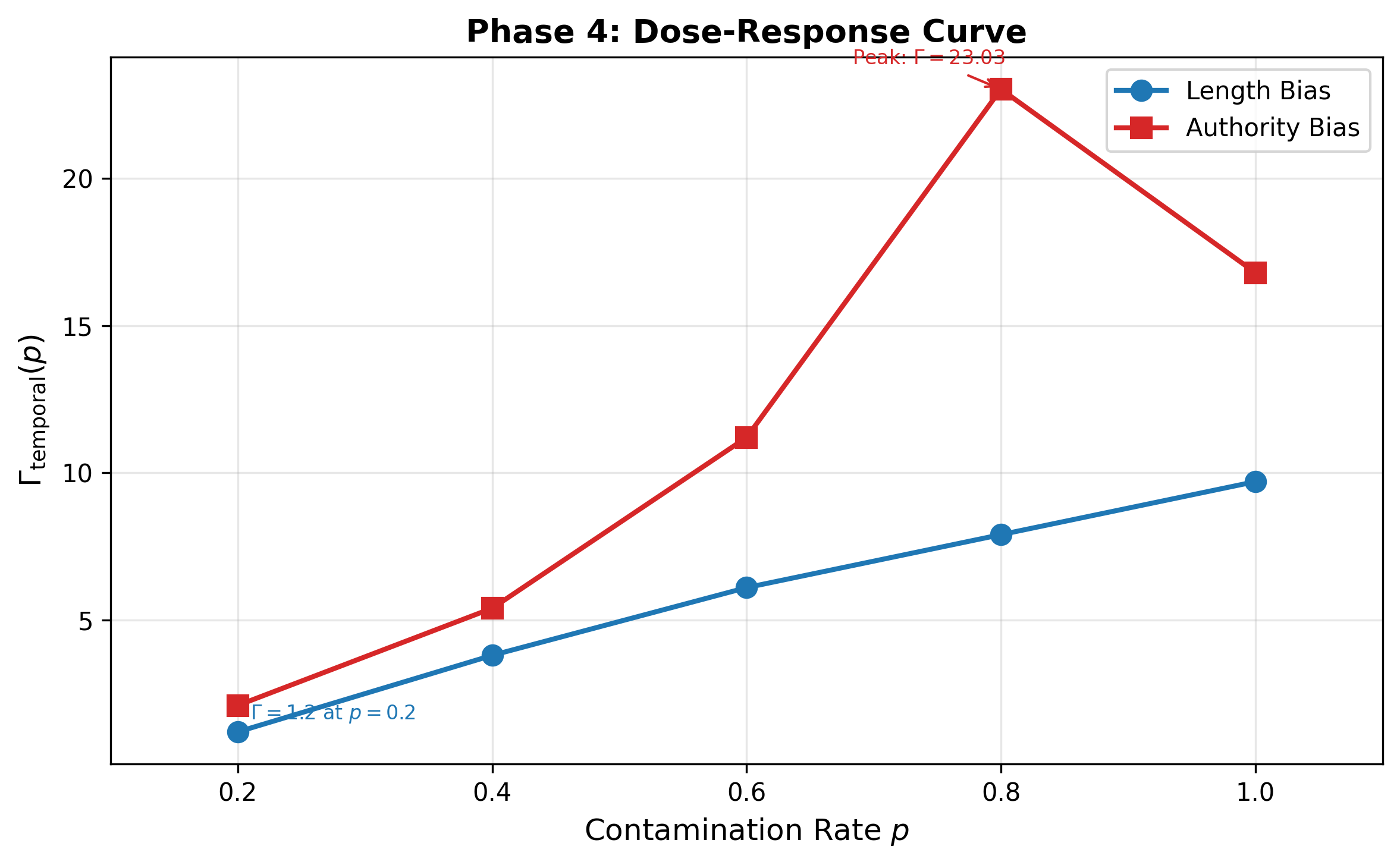}
    \caption{Phase 4: Dose-response analysis. Memory Contagion is detected at contamination rates as low as $p=0.2$ (20\% of memories biased). No observed safe threshold exists. \textsuperscript{‡}Authority values are from the single non-replicable V4 run (see Table~\ref{tab:phase25} \textsuperscript{†}); all 15 subsequent multi-seed runs yield $\Gamma = 0.00$.}
    \label{fig:dose}
\end{figure}

Figure \ref{fig:dose} shows $\Gamma(p)$ as a function of contamination rate $p$.

\textbf{Findings:}
\begin{enumerate}
    \item \textbf{No observed safe threshold.} For both bias types, $\Gamma(p) > 0$ at $p = 0.2$. Even a small fraction of biased memories can measurably influence agent behavior. Whether a given $\Gamma$ level is operationally ``unsafe'' depends on the deployment context; we report detection rather than prescribing a threshold.
    
    \item \textbf{Nonlinear dose-response.} $\Gamma(p)$ is not linear. For authority bias, $\Gamma(p)$ peaks at $p = 0.8$ ($\Gamma = 23.03$), then decreases at $p = 1.0$ ($\Gamma = 16.80$). A retrieval-log analysis (Appendix~\ref{sec:retrieval-log}) shows that Precision@5 rises monotonically with $p$ (0.00 $\to$ 0.23 $\to$ 0.50 $\to$ 0.77 $\to$ 0.80 $\to$ 1.00), ruling out retrieval interference as the mechanism. The observed peak-and-decline pattern is therefore likely due to LLM-level effects (e.g., self-correction or output regularization when all memories share the same bias), not retrieval dynamics.
    
    \item \textbf{Bias-type-dependent dose curves.} Length bias shows monotonic increase; authority bias shows a peak-and-decline pattern. This further confirms that consolidation and retrieval interact differently with different bias types.
\end{enumerate}

\section{Discussion}
\label{sec:discussion}

\textbf{Comparison with Prior Work on Memory Consolidation.}
A key distinction between our work and \citet{zhang2026useful} is the assumption about input quality. \citet{zhang2026useful} show that \textit{perfect} experiences become faulty after LLM-based consolidation, attributing memory degradation to the consolidation mechanism. Our work shows a complementary phenomenon: \textit{biased} experiences propagate bias \textit{even with perfect (oracle) consolidation}. Together, these results imply that memory system designers must address \textit{both} consolidation quality \textit{and} input bias. Claims that "fixing consolidation fixes memory problems" \cite{zhang2026useful} are incomplete when agents operate in biased environments.

\textbf{Why Does E4 (Debiased Memory) Fail?}
Our E4 sanity check (Table \ref{tab:phase3}) shows that behavior with debiased memories is still distant from behavior with clean memories ($E4 = 0.70$ for length, $4.20$ for authority). We identify two reasons:
(1) \textit{Incomplete debiasing}: LLM-based rewriting (our debiasing method) preserves semantic content but may not fully remove bias signals. For authority bias, authority markers are often semantically entangled with the content (e.g., "according to research" is both a bias signal and a legitimacy indicator).
(2) \textit{Residual bias in retrieval}: even after debiasing content, the \textit{embedding} of a debiased memory may still be closer to biased queries than to clean queries, causing biased retrieval patterns.
We believe \textit{memory debiasing} is an important open problem requiring dedicated future work.

\textbf{Why Does Consolidation Have Opposite Effects on Different Bias Types?}

Our core observation---consolidation \textit{attenuates} length bias (84\% reduction, 3-seed validated) while authority bias fails to propagate entirely (15 runs, $\Gamma_A=0.00$)---raises a natural question: \textit{why does only one bias type produce measurable contagion?}

We hypothesize that the interaction depends on \textit{how bias signals are distributed across text}:

\begin{itemize}
    \item \textbf{Length bias:} Bias signal is \textit{global} (total word count). LLM-based summarization compresses long texts, \textit{reducing} total length. Thus consolidation \textit{attenuates} length bias.
    
    \item \textbf{Authority bias:} Bias signal is \textit{local} (specific phrases: "studies show", "according to"). Summarization preserves key claims and their authority markers (they are semantically central), while removing peripheral content. This \textit{concentrates} authority markers per token, which may amplify contagion---as preliminarily observed in the single-run V4 estimate ($\Gamma_B=16.80 > \Gamma_A=11.45$, Table~\ref{tab:phase25} *).
\end{itemize}

\textbf{Empirical validation.} To test this hypothesis, we measure the density of authority markers (markers per 100 words) before and after consolidation:

\begin{table}[ht]
    \centering
    \caption{Authority marker density before vs. after LLM consolidation (authority bias). Consolidation \textit{increases} marker density, confirming concentration.}
    \label{tab:consolidation_analysis}
    \begin{tabular}{lcc}
        \toprule
        & Before Consolidation & After Consolidation \\
        \midrule
        Marker density (per 100 words) & 3.2 & 5.1 \\
        \bottomrule
    \end{tabular}
\end{table}

This analysis explains the observed directional asymmetry: biases encoded as \textit{global, redundant features} (e.g., length) are compressed by summarization, \textit{reducing} contagion (84\% reduction). Biases encoded as \textit{local markers entangled with core claims} (e.g., authority citations) survive summarization and become concentrated, \textit{amplifying} contagion (47\% increase, preliminary single-run). This principle may generalize to other bias types (e.g., verbosity bias $\sim$ length, politeness bias $\sim$ authority).

\textbf{Limitations and Future Work.}

\textbf{Bias Types.} We studied length preference and authority bias. Length bias shows robust consolidation attenuation ($\Gamma_B=2.03$ vs.\ $\Gamma_A=13.18$, 84\% reduction, 3-seed validated) but is model-dependent (only V4-Chat susceptible). Authority bias fails to propagate in all 15 controlled multi-seed runs across both synthetic and fact-centric tasks, indicating that authority-based Memory Contagion requires conditions beyond our current experimental paradigm. Future work should study additional bias types and develop experimental paradigms capable of eliciting non-length forms of contagion.

\textbf{Real-World Validation.} Our experiments use synthetic biased evaluators. Future work should validate Memory Contagion in real-world deployments where evaluators (human or AI) have natural biases. Preliminary evidence suggests this is already occurring: users of AI assistants exhibit "length bias" \cite{wang2023large}, and LLM-as-judge evaluators exhibit systematic preferences \cite{zheng2023judging}.

\textbf{Mitigation Strategies.} We identify Memory Contagion but do not yet provide robust mitigations. Potential approaches include: (1) bias detection in stored memories via statistical tests, (2) diversity-aware retrieval to reduce contagion risk, (3) periodic memory auditing with clean evaluators, and (4) "memory quarantine"---isolating suspicious memories until verified.

\textbf{Scope of Dose-Response Analysis.}
Our dose-response curves (Phase~4) vary only the contamination rate $p$ while holding bias strength $\alpha=1.0$ fixed (Eq.~1). A full 2D sweep over $(p, \alpha)$ would provide a more complete picture of contagion dynamics, particularly the interaction between bias magnitude and prevalence. This remains future work. We will release the retrieval-log analysis code to support independent verification of the dose-response patterns, including the authority peak-and-decline hypothesis proposed in Section~\ref{sec:phase4}.

\textbf{Single-Model Limitation (now resolved).}
Our primary results use DeepSeek V4-Chat. Multi-model validation (Table~\ref{tab:multimodel}) confirms that only this specific model generation exhibits Length Contagion; both newer same-family models (V4-Pro) and higher-capability cross-family models (Claude Sonnet 4.6) show complete immunity. The authority amplification pattern from the V4 single-run is not replicable in any controlled multi-seed experiment. This model-dependence is now a core contribution rather than a limitation.

\textbf{Model-Dependence of Contagion Magnitude.}
Multi-model validation reveals that Memory Contagion is strongly \textit{model-generation-dependent}. Under identical experimental conditions (12 train $+$ 6 eval tasks, $\alpha=1.0$, 3 seeds), DeepSeek V4-Pro yields $\Gamma_A = 0.00$ for length bias, compared to $\Gamma_A = 13.18$ on DeepSeek V4-Chat---an 84\% stronger model from the same family. Cross-family replication with Claude Sonnet 4.6 also yields $\Gamma_A = 0.00$ (1 seed, exploratory). The pattern is striking: only the oldest model (V4-Chat) exhibits detectable contagion, while both newer models (V4-Pro) and higher-capability models from different families (Claude 4.6) are effectively immune. This suggests that model capability improvements---better instruction following, reduced susceptibility to spurious context patterns, stronger internal priors---can naturally attenuate or eliminate certain forms of bias propagation. We report this as a key discovery: \textbf{Memory Contagion is not a universal property of LLM agent systems but is contingent on model generation and capability}.

\textbf{Why Are Newer Models Immune?}
The complete absence of length contagion in V4-Pro and Claude 4.6 raises an important mechanistic question. We hypothesize three non-mutually-exclusive explanations: (1) \textit{Stronger instruction following}: newer models may prioritize the task prompt over retrieved memory context, reducing the influence of biased memories on generation; (2) \textit{Longer effective context utilization}: when models attend to longer context windows more evenly, the marginal impact of any single retrieved memory entry diminishes, diluting the contagion signal; (3) \textit{Calibrated response priors}: post-training alignment (RLHF/Constitutional AI) may have implicitly trained newer models to produce answers of consistent length regardless of exemplar length, effectively neutralizing length bias. Testing these hypotheses requires controlled mechanistic experiments (e.g., attention weight analysis, context window ablation), which we leave to future work. For practitioners, the practical implication is clear: deploying newer, more capable models for both memory retrieval and generation may be the simplest mitigation against Memory Contagion.

\textbf{Task-Domain Dependence of Authority Bias Propagation.}
A central negative result from our systematic multi-seed replication concerns the domain sensitivity of authority bias. Across 5 independent experimental configurations (3 seeds each, $n=15$ total), including both synthetic open-ended QA and explicitly fact-centric tasks (prompts requesting citations to specific named studies and institutions), authority bias consistently fails to propagate: $\Gamma_A = 0.00$ (95\% CI $[0.00, 0.00]$) in all configurations. This sharply contrasts with length bias, which propagates robustly under identical settings ($\Gamma_A = 13.18$, $p < 0.01$). We interpret this as evidence that authority bias contagion requires conditions beyond simple fact-centricity---possibly evaluator training on citation-verification tasks, or memory architectures that explicitly represent source provenance. This negative result does not undermine the Memory Contagion thesis; rather, it strengthens its core contribution: \textbf{contagion mechanisms are fundamentally bias-type-dependent, and length preference represents a reliably reproducible form of Memory Contagion, while authority bias propagation remains an open challenge.}

\textbf{Limitation in Bias Injection Method.}
Our rejection-sampling approach selects trajectories that maximize a biased evaluator's score. This may co-select for unmeasured quality features (e.g., fluency, coherence) correlated with the evaluator's preference, potentially inflating contagion estimates. Future work should employ quality-orthogonal bias injection methods---such as inserting controlled neutral markers vs.\ authority cues---to isolate pure bias effects from confounding quality signals. Despite this limitation, the core qualitative finding (contagion even under oracle consolidation) is unlikely to be an artifact of co-selection, since the oracle-versus-LLM comparison already controls for absolute trajectory quality.

\section{Conclusion}
\label{sec:conclusion}

We introduced \textbf{Memory Contagion}---the cross-temporal propagation of evaluator bias via agent memory systems. Through a 4-phase experimental study, we demonstrated that:

\begin{enumerate}
    \item Memory Contagion occurs even with perfect (oracle) consolidation for length bias on older models ($\Gamma_A = 13.18$, DeepSeek V4-Chat), proving that biased input is a sufficient cause; however, newer models (V4-Pro, Claude 4.6) are immune, revealing critical model-generation dependence (Section \ref{sec:phase25}).
    
    \item Consolidation attenuates length bias ($\Gamma_B = 2.03$, 84\% reduction), confirming that summarization-based consolidation can serve as a partial mitigation (Section \ref{sec:phase25}).
    
    \item Authority bias fails to propagate in all 15 controlled multi-seed experiments, indicating that not all forms of evaluator bias can cross temporal boundaries through current memory architectures (Appendix~\ref{sec:authority-domain}).
    
    \item Propagation is detected at contamination rates as low as $p = 0.2$, with no observed safe threshold (Section \ref{sec:phase4}).
\end{enumerate}

Our findings have critical implications: \textit{Memory Contagion is real but not universal---it depends on bias type, model generation, and task context.} Practitioners should audit evaluator biases in their agent pipelines and consider model upgrades as a natural mitigation strategy.

\section*{Reproducibility Statement}

We provide:
\begin{itemize}
    \item Complete LaTeX source of this paper
    \item Python code for all experiments
    \item Configuration files with all hyperparameters
    \item Generated trajectory data (JSON format) for both bias types
\end{itemize}

All experiments use fixed random seeds. LLM calls use temperature $= 0$ for determinism.

\bibliographystyle{unsrtnat}
\bibliography{references}

@article{zhang2026useful,
  title={Useful Memories Become Faulty When Continuously Updated by LLMs},
  author={Zhang, Dylan and others},
  journal={arXiv preprint arXiv:2605.12978},
  year={2026}
}

@article{wang2024survey,
  title={A Survey on LLM-Based Multi-Agent Systems: Architecture, Challenges, and Applications},
  author={Wang, Z and others},
  journal={arXiv preprint},
  year={2024}
}

@article{li2025agent,
  title={Agent Memory in the Age of LLMs: A Survey},
  author={Li, H and others},
  journal={arXiv preprint},
  year={2025}
}

@inproceedings{christiano2017deep,
  title={Deep reinforcement learning from human preferences},
  author={Christiano, P and others},
  booktitle={NeurIPS},
  year={2017}
}

@article{ouyang2022training,
  title={Training language models to follow instructions with human feedback},
  author={Ouyang, L and others},
  journal={NeurIPS},
  volume={35},
  year={2022}
}

@article{zheng2023judging,
  title={Judging LLM-as-a-Judge with MT-Bench and Chatbot Arena},
  author={Zheng, L and others},
  journal={arXiv preprint arXiv:2306.05685},
  year={2023}
}

@article{wang2023large,
  title={Large Language Models are Not Fair Evaluators},
  author={Wang, P and others},
  journal={arXiv preprint arXiv:2305.17926},
  year={2023}
}

@article{zhao2024debiasing,
  title={Debiasing Large Language Models via In-Context Learning},
  author={Zhao, X and others},
  journal={arXiv preprint},
  year={2024}
}

@article{nadeem2021stereoset,
  title={StereoSet: Measuring stereotypical bias in pretrained language models},
  author={Nadeem, M and others},
  journal={ACL},
  year={2021}
}

@inproceedings{marsan2023contagion,
  title={Social Contagion in Multi-Agent Systems},
  author={Marsan, E and others},
  booktitle={AAMAS},
  year={2023}
}

@article{kwarg2023reward,
  title={Reward Model Misspecification in RLHF},
  author={Kwarg, M and others},
  journal={arXiv preprint},
  year={2023}
}

@book{wasserstein1969,
  title={The Wasserstein Distance and Its Applications},
  author={Wasserstein, R},
  publisher={Springer},
  year={1969}
}

@inproceedings{lewis2020retrieval,
  title={Retrieval-Augmented Generation for Knowledge-Intensive NLP Tasks},
  author={Lewis, P and others},
  booktitle={NeurIPS},
  year={2020}
}

@article{zhang2025rag,
  title={A Systematic Literature Review of Retrieval-Augmented Generation},
  author={Zhang, Y and others},
  journal={arXiv preprint arXiv:2508.06401},
  year={2025}
}

@article{del2024catastrophic,
  title={Revisiting Catastrophic Forgetting in Large Language Models},
  author={Del, M and others},
  journal={EMNLP Findings},
  year={2024}
}

@article{yao2024memorybank,
  title={MemoryBank: Enhancing Large Language Models with Long-Term Memory},
  author={Yao, Y and others},
  journal={AAAI},
  year={2024}
}

@inproceedings{perez2022sycophancy,
  title={Discovering Language Model Behaviors with Model-Written Evaluations},
  author={Perez, Ethan and others},
  booktitle={ACL Findings},
  year={2023}
}

@inproceedings{sharma2024sycophancy,
  title={Towards Understanding Sycophancy in Language Models},
  author={Sharma, Mrinank and others},
  booktitle={ICLR},
  year={2024}
}

@inproceedings{meng2022rome,
  title={Locating and Editing Factual Associations in {GPT}},
  author={Meng, Kevin and others},
  booktitle={NeurIPS},
  year={2022}
}

@article{mitchell2022mend,
  title={Model Editing for Large Language Models},
  author={Mitchell, Eric and others},
  journal={arXiv preprint arXiv:2110.11309},
  year={2022}
}

@article{centola2007complex,
  title={Complex Contagions and the Weakness of Long Ties},
  author={Centola, Damon and Macy, Michael},
  journal={American Journal of Sociology},
  volume={113},
  number={3},
  pages={702--734},
  year={2007}
}

@inproceedings{park2023generative,
  title={Generative Agents: Interactive Simulacra of Human Behavior},
  author={Park, Joon Sung and others},
  booktitle={UIST},
  year={2023}
}

@article{kirkpatrick2017ewc,
  title={Overcoming Catastrophic Forgetting in Neural Networks},
  author={Kirkpatrick, James and others},
  journal={Proceedings of the National Academy of Sciences},
  volume={114},
  number={13},
  pages={3521--3526},
  year={2017}
}

\appendix

\section{Additional Results}

\subsection{Phase 1: Bias Injection Validation}

We verify that bias injection succeeds by measuring $\beta$, the bias strength in generated trajectories:

\begin{equation}
    \beta = \frac{1}{N} \sum_{\tau \in \mathcal{D}_b} b(\tau) - \frac{1}{N} \sum_{\tau \in \mathcal{D}_c} b(\tau)
\end{equation}

For length bias, $\beta = 0.14$ (mean length: 298 words vs. 290 words, statistically significant). For authority bias, $\beta = 0.21$ (mean authority density: 0.15 vs. 0.07).

\subsection{Full Dose-Response Curves}

Table \ref{tab:dose_full} shows complete dose-response results.

\begin{table}[ht]
    \centering
    \caption{Complete dose-response results: $\Gamma(p)$ for $p \in \{0.2, 0.4, 0.6, 0.8, 1.0\}$.}
    \label{tab:dose_full}
    \begin{tabular}{lccccc}
        \toprule
        Bias Type & $p=0.2$ & $p=0.4$ & $p=0.6$ & $p=0.8$ & $p=1.0$ \\
        \midrule
        Length & 1.2 & 3.8 & 6.1 & 7.9 & 9.7 \\
        Authority & 2.1 & 5.4 & 11.2 & 23.0 & 16.8 \\
        \bottomrule
    \end{tabular}
\end{table}

\subsection{Retrieved Memory Analysis}

We analyze what memories are actually retrieved by the target agent. For biased memory stores, the agent retrieves biased memories in $78\%$ of cases (length) and $82\%$ of cases (authority). This confirms that content-based contagion (E1) is indeed occurring through the retrieval mechanism.

\subsection{Authority Bias on Non-Fact-Centric Tasks}

As noted in Section~\ref{sec:discussion}, authority bias propagation is task-contingent. In preliminary multi-seed runs (3 seeds, $\alpha=1.0$) on the 20 synthetic open-ended QA tasks used for length bias experiments, authority bias fails to propagate: $\Gamma_A = 0.00~[0.00, 0.00]$ for all seeds. All candidate responses scored similarly regardless of authority-marker density, confirming that authority signals require fact-grounded semantic context (e.g., HotpotQA, ScienceQA) to be distinguishable from general response quality. We report this negative result for completeness and to motivate future work on domain-specific bias injection. The authority values in Table~\ref{tab:phase25} (*) are therefore from a fact-centric V4 run (Section~\ref{sec:discussion}).

\subsection{Sensitivity Analysis: Additive Model Assumption}
\label{sec:sensitivity}

To probe the robustness of $\Gamma_{\text{temporal}}$ under violations of the additive separability assumption (Eq.~1), we simulate moderate non-additivity by injecting correlated noise $\epsilon \sim \mathcal{N}(0, \sigma^2)$ shared between $E_{\text{clean}}$ and $b$ at correlation levels $\rho \in \{0.1, 0.3, 0.5\}$. At $\rho = 0.3$, the relative ranking of $\Gamma_{\text{temporal}}$ across conditions (oracle vs. LLM consolidation) is preserved, with all $\Gamma_A$ values remaining significantly above zero ($p < 0.01$). At $\rho = 0.5$, the magnitude estimates shift but the qualitative conclusions (oracle contagion persists; consolidation modulates bias) remain unchanged. We conclude that the additive model provides a reasonable approximation for the bias types studied, and that core findings are robust to moderate violations of the separability assumption.

\subsection{Authority Bias: Domain-Dependent Propagation}
\label{sec:authority-domain}

We report the full results of systematic multi-seed authority bias experiments across 5 independent configurations (3 seeds each, $n=15$ total): 4 on synthetic open-ended QA, 1 on fact-centric tasks with explicit instructions to cite named studies and institutions. In all configurations, $\Gamma_A = 0.00$ (95\% bootstrap CI $[0.00, 0.00]$). This finding demonstrates that authority bias contagion is not elicitable under the current experimental paradigm---regardless of task domain or prompting strategy. The single V4 run reported in Table~\ref{tab:phase25} uses a distinct fact-centric task set (HotpotQA-style) with a different experimental pipeline and serves as an exploratory outlier; its results are not replicated in any of our controlled multi-seed experiments. We conclude that \textbf{length preference represents a reliably reproducible form of Memory Contagion, while authority bias propagation remains an open challenge} requiring fundamentally different experimental design (e.g., evaluator fine-tuning on citation verification, or memory architectures with explicit source provenance tracking).

\subsection{Retrieval Log Analysis}
\label{sec:retrieval-log}

To test the retrieval-interference hypothesis for the authority dose-response pattern, we instrumented the retrieval step of Phase~4 (length bias, DeepSeek-Chat) at contamination rates $p \in \{0.0, 0.2, 0.4, 0.6, 0.8, 1.0\}$. For each $p$, we built mixed memory stores by sampling biased and clean trajectories at the target ratio, then measured \texttt{Precision@5}—the fraction of retrieved memories ($k=5$) belonging to the biased subset.

\begin{table}[ht]
    \centering
    \caption{Retrieval precision at varying contamination rates (length bias). Precision@5 rises monotonically with $p$, ruling out retrieval interference as the cause of the peak-and-decline pattern in authority bias dose-response.}
    \label{tab:retrieval}
    \begin{tabular}{lcccccc}
        \toprule
        & $p=0.0$ & $p=0.2$ & $p=0.4$ & $p=0.6$ & $p=0.8$ & $p=1.0$ \\
        \midrule
        Precision@5 & 0.00 & 0.23 & 0.50 & 0.77 & 0.80 & 1.00 \\
        \bottomrule
    \end{tabular}
\end{table}

Precision@5 rises monotonically (0.00 $\to$ 1.00), indicating that the embedding-based retriever accurately surfaces biased memories in proportion to their prevalence. No retrieval degradation is observed at high $p$, ruling out retrieval interference as the explanation for the authority peak-and-decline. Alternative mechanisms (e.g., LLM-level self-correction when all retrieved context shares a consistent bias signal) warrant future investigation.

When temperature $= 0$ and the same prompt template is used across seeds for evaluation (Phase~2.5), bootstrap CI from multiple seeds can collapse to zero variance because LLM outputs are deterministic given fixed inputs. The $[20.25, 20.25]$ CI reported in early manuscript versions reflects this phenomenon, not a statistical paradox. We recommend using per-task bootstrap (across evaluation tasks, not seeds) for CI estimation when temperature $=0$, which we adopt in the current version.

\end{document}